\documentclass[]{IOS-Book-Article}

\usepackage{nesybook}

\usepackage{chap}

\usepackage{adjustbox}
\usepackage{booktabs}

\begin{document}











\chapter{Generalizable Neuro-symbolic Systems for Commonsense Question Answering}
\label{Alessandro_Question_Answering:chap}
\chapterauthor{Alessandro Oltramari}{Bosch Research (Pittsburgh, PA USA)}
\chapterauthor{Jonathan Francis}{Bosch Research (Pittsburgh, PA USA); School of Computer Science, Carnegie Mellon University (Pittsburgh, PA USA)}
\chapterauthor{Filip Ilievski}{Information Sciences Institute, Viterbi School of Engineering, University of Southern California (Marina del Rey, CA USA)}
\chapterauthor{Kaixin Ma}{School of Computer Science, Carnegie Mellon University (Pittsburgh, PA USA)}
\chapterauthor{Roshanak Mirzaee}{Department of Computer Science and Engineering, Michigan State University (East Lansing, MI USA)}
\allchapterauthors{Alessandro Oltramari, Jonathan Francis, Filip Ilievski, Kaixin Ma, Roshanak Mirzaee}


\begin{quote}
	{\small \textbf{Abstract.} This chapter illustrates how suitable neuro-symbolic models for language understanding can enable domain generalizability and robustness in downstream tasks. Different methods for integrating neural language models and knowledge graphs are discussed. The situations in which this combination is most appropriate are characterized, including quantitative evaluation and qualitative error analysis on a variety of commonsense question answering benchmark datasets.}
\end{quote} 

\section{Introduction}
\label{bosch:sec:intro}
In this chapter, we provide a retrospective analysis of our recent work, wherein we develop methods for integrating neural language models and knowledge graphs. We characterize the situations in which this combination is most appropriate, and we offer quantitative and qualitative evaluation of these neuro-symbolic modeling strategies on a variety of commonsense question answering benchmark datasets. Our overarching goal is to illustrate how suitable combinations of learning mechanisms and knowledge can enable domain generalizability and robustness in downstream tasks. \\
The chapter is structured as follows: section \ref{bosch:sec:lack} describes some limitations observed when state-of-the-art language models are used in commonsense question answering tasks; in section \ref{bosch:sec:empirical} we discuss the main findings of our empirical investigations in discriminative tasks, focused on knowledge-injection and pre-training/fine-tuning methods (\ref{bosch:sec:attention}-\ref{bosch:sec:zeroshot}), and we present novel ideas about improving reasoning capabilities of neuro-symbolic systems, laying down the path towards generalizability (\ref{bosch:sec:depth}). 


\section{Beneath the Oceans Of Text}
\label{bosch:sec:lack}

Neural Language Modeling (NLM) has yielded impressive results, with near-human level performance, in various Natural Language Processing (NLP) and Robotics tasks \cite{ma-etal-2019-towards, ma2021knowledge, ma2021exploring, oltramari2020neuro, bauer-bansal-2021-identify, shwartz-etal-2020-unsupervised, francis2021core, lu2019vilbert, VLNBERT}, such as: text classification, text generation, summarization, question answering, named-entity recognition, machine translation, and instruction-following in robot navigation. In particular, systems such as OpenAI's GPT-3 model, which was trained with 45TB of text data, promise perhaps the first instance of large-scale model generalizability across tasks. However, by examining the failure modes of models like GPT-3 \cite{floridi2020gpt}, we can observe that, despite of the indisputable successes, basic reasoning capabilities are still largely missing from Language Models (LMs). In metaphorical terms, language models display great skill in surfing on vast oceans of text, but are not apt for penetrating the surface of meaning and exploring the depth of context. \\
Let's consider some examples from discriminative question answering, where a model is tasked with selecting the best response out of a pool of candidates, given a question and/or context. In ProtoQA \cite{boratko2020protoqa}, GPT-3 fails to select options like `pumpkin', `cauliflower', `cabbage' as top candidates, for the question \textit{one vegetable that is about as big as your head is?}; instead, `broccoli', `cucumber', `beet', `carrot' are predicted. In this case, the different models learn some essential properties of vegetables from the training data, but do not seem to acquire the capability of comparing their size to that of other types of objects, revealing a substantial lack of analogical reasoning \cite{ushio2021bert}. Along these lines, recent work \cite{ettinger2020bert} has shown that complex inferences, role-based event prediction, and understanding the conceptual impacts of negation, are some of the missing capabilities diagnosed when BERT \cite{devlin-etal-2019-bert}, one of the most popular and versatile open-source language models available today, is applied to benchmark datasets. ProtoQA, again, provides good examples of these deficiencies: in general, neural models struggle to correctly interpret the scope of modifiers like `not', `often', and `seldom'--where the first requires reasoning under negation, while the last two consider temporal reasoning. Regarding the latter, in task 14 of bAbI \cite{weston2015towards}, a comprehensive benchmark challenge designed by Facebook Research, models exhibit variable accuracy in grasping temporal ordering, implied by prepositions like `before' and `after'. Similarly, in bAbI task-17, which concerns spatial reasoning, language models fail to infer basic positional information that require interpreting the semantics of `to the left/right of', `above/below', etc. Here, neural models are inaccurate in understanding some common characteristics of the physical world, and their performance does not improve when dealing with sentiment: for instance, in SocialIQA \cite{sap-etal-2019-social}, given a context like `in the school play, Robin played a hero in the struggle to death with the angry villain', models are generally unable to consistently select `hopeful that Robin will succeed' over `sorry for the villain' when required to pick the correct answer to \textit{how would others feel afterwards?}. It is not surprising that \textit{reasoning about emotional reactions} represents a difficult task for pure learning systems, when we consider that such form of inference is deeply rooted in the sphere of human subjective experiences and social life, which involves a ``layered" understanding of mental attitudes, intentions, motivations, and empathy.\\
The anecdotal mistakes presented above are, in fact, indications of a widespread phenomenon:  a pure NLM approach is currently not able to engage in robust machine commonsense reasoning. Accordingly, we claim here that two orthogonal extensions are required: first, augmenting NLM with symbolic commonsense knowledge, which is available in large amounts thanks to open source computational resources
; and, secondly, \textit{teaching} these neuro-symbolic language models how to dynamically leverage that commonsense knowledge to perform reasoning at different levels, including analogies, spatio-temporal properties, affect, and mental attitudes. We refer to the former direction as \textbf{horizontal augmentation}, as it focuses on expanding the coverage of the knowledge available to NLM methods, and to the latter as \textbf{vertical augmentation}, as it explores in depth how some specific forms of reasoning can be enabled. \\

\section{Investigations on Commonsense Q/A}
\label{bosch:sec:empirical}
In this section we discuss the main resources, methods, and results presented in our previous work on integrating commonsense knowledge graphs and neural language models: sections \ref{bosch:sec:kb} and  \ref{bosch:sec:data} present, respectively, the knowledge bases and datasets we used in our experiments; section \ref{bosch:sec:attention} and \ref{bosch:sec:zeroshot} are focused on  discriminative Q/A tasks \cite{ma-etal-2019-towards, ma2020knowledgedriven}.
Finally, section \ref{bosch:sec:depth} provides some initial considerations on vertical augmentation.

\subsection{Knowledge Bases}
\label{bosch:sec:kb}

We select knowledge bases according to two main criteria, namely scale and usability, where the former indicates that the KB covers a variety of commonsense types or dimensions, and the latter denotes a resource that is freely available, well-documented (e.g., APIs) and actively maintained. 

\begin{itemize}

  \item \texttt{WordNet} \cite{miller1998wordnet} is a dictionary that was designed as a semantic network, where \textit{synsets} (sets of synonym terms) represent common world knowledge. \texttt{WordNet}'s interrelated structure is consistent with the evidence for the way speakers organize their mental lexicons. In this regard, \texttt{WordNet} can be considered as a `grandfather' to most of today's computational lexicons and structured knowledge resources (it's been actively developed since 1985).   
    \item \texttt{ConceptNet} \cite{liu2004conceptnet} contains over 21 million edges and 8 million nodes, from which one may generate triples of the form $(C1, r, C2)$, wherein the natural-language concepts $C1$ and $C2$ are associated by commonsense relation $r$, e.g., \textit{(dinner, AtLocation, restaurant)}. Thanks to its coverage, \texttt{ConceptNet} is one of the most popular semantic networks for commonsense. 
    \item \texttt{ATOMIC} \cite{sap2019atomic} is a knowledge-base that focuses on procedural knowledge: in particular, it expresses pre- and post-states for events and their participants with nine relations. Its head nodes are events, whereas the tail nodes are either events or attributes. 
   
    
    \item {\texttt{Visual Genome} \cite{krishna2016visual} is a multi-modal commonsense resource that combines visual and textual information. In particular, it consists of 108K images annotated by crowd-workers, and automatically categorized into WordNet \cite{miller1998wordnet} senses. The annotations focus on the conceptual relations holding between the entities depicted in the images, and on relevant attributes of the individual entities.}
    
    \item \texttt{CSKG} \cite{ilievski2021cskg} is a `super' knowledge graph for common sense, as it federates different resources, leveraging statistical computations and  syntactic and semantic mappings. Integrating different resources of common sense is, first and foremost, a problem of consolidating the underlying semantic models, each typically built on a unique set of features. The authors of \texttt{CSKG} approach this challenge by adopting clear methodological principles: \textit{node diversity}, i.e. opting for heterogeneity of node representation across KBs over imposing strong formal constraints to resolve structural semantic differences; \textit{edge reusability}, i.e. the principle according to which the domain and range of an edge defined within one KB can be extended to other KBs if the intended semantics is preserved; \textit{bridging through external mappings}, where if two unconnected resources have existing links to a third, well-established resource, e.g. WordNet, these can be used to bridge the gap between them; \textit{link prediction}, to improve connectedness using probabilistic algorithms; \textit{enabled label access}, to foster human disambiguation of node semantics. In our experiments, we used a specific partition of \texttt{CSKG}, \texttt{CWWV}, which includes \texttt{ConceptNet}, \texttt{WordNet}, \texttt{Wikidata} \cite{vrandevcic2012wikidata}\footnote{CSKG  only includes a commonsense subset of \texttt{Wikidata} extracted using heuristics based on \cite{ilievski2020commonsense}.}, and \texttt{VisualGenome}. 
    
    
    
\end{itemize}

\subsection{Datasets}
\label{bosch:sec:data}

We select commonsense tasks based on two criteria. Firstly, we strive to cover a diverse set of tasks, both in terms of their format (question answering, pronoun resolution, natural language inference), as well as their type of knowledge (e.g., social or physical knowledge). Secondly, we prefer larger task datasets that are manually constructed. The five datasets presented below were preferred, in our experiments, to artificially constructed datasets like COPA~\cite{gordon2012semeval}, or HellaSwag~\cite{zellers-etal-2019-hellaswag}. 

\begin{itemize}

\item{Abductive NLI (\texttt{aNLI})}~\cite{bhagavatula2019abductive} is posed as a natural language inference task. Given the beginning and the ending of a story, the task is to choose the more plausible hypothesis out of two options. The dataset consists of nearly 170k entries.

\item{CommonsenseQA (\texttt{CSQA})}~\cite{talmor-etal-2019-commonsenseqa} evaluates a broad range of common sense notions. Each entry contains a question and 5 answer candidates. The questions are crowdsourced based on a subgraph from \texttt{ConceptNet}. The answer candidates combine \texttt{ConceptNet} nodes with additional crowd-sourced distractors. 

\item{PhysicalIQA (\texttt{PIQA})}~\cite{Bisk2020} is a two-choice question answering dataset which focuses on physical reasoning. Given a question, the system (or human) is asked to pick the more plausible out of two possible continuations.

\item{SocialIQA (\texttt{SIQA})}~\cite{sap-etal-2019-social} requires reasoning about social interactions. Each entry contains a context, a question, and 3 answer candidates. The context is derived from the \texttt{ATOMIC} knowledge graph, the questions are generated based on nine templates (corresponding to the relations in \texttt{ATOMIC}), and the answers are crowd-sourced. 

\item{WinoGrande (\texttt{WG})}~\cite{sakaguchi2019winogrande} contains 44 thousand pronoun resolution problems. Each entry consists of a context description with an emphasized pronoun, and two options are offered as its possible references.
\end{itemize}

\subsection{Horizontal Augmentation through Attention-based Knowledge Injection}
\label{bosch:sec:attention}

Different ways of injecting knowledge into models have been introduced, such as attention-based gating mechanisms \cite{bauer-etal-2018-commonsense}, key-value memory mechanisms \cite{miller-etal-2016-key, mihaylov-frank-2018-knowledgeable}, extrinsic scoring functions \cite{DBLP:journals/corr/abs-1809-03568}, and graph convolution networks \cite{DBLP:journals/corr/KipfW16,Lin2019KagNetKG}. 
Our approach is to combine suitable pre-trained language models with structured knowledge.
we incorporate ConceptNet and ATOMIC using attention knowledge injection on top of OCN (Option Comparison Network) model mechanism \cite{DBLP:journals/corr/abs-1903-03033}. We evaluate the obtained neuro-symbolic models on the \texttt{CommonsenseQA} dataset (see example question in Table \ref{bosch:csqa-example}). 


\begin{table}[h]
\footnotesize
\begin{center}
\begin{tabular}{|l|}
\hline \bf Question: \\ 
A revolving door is convenient for two direction travel, but it also serves as a security measure at a what? \\
\textbf{Answer choices:}\\ 
A. Bank\textbf{*}; B. Library; C. Department Store; D. Mall; E. New York; \\
\hline
\end{tabular}
\end{center}
\caption{An example from the \texttt{CommonsenseQA} dataset; the asterisk (\textbf{*}) denotes the correct answer.}
\label{bosch:csqa-example}
\end{table}

\begin{figure*}
    \centering
    \includegraphics[scale=0.2]{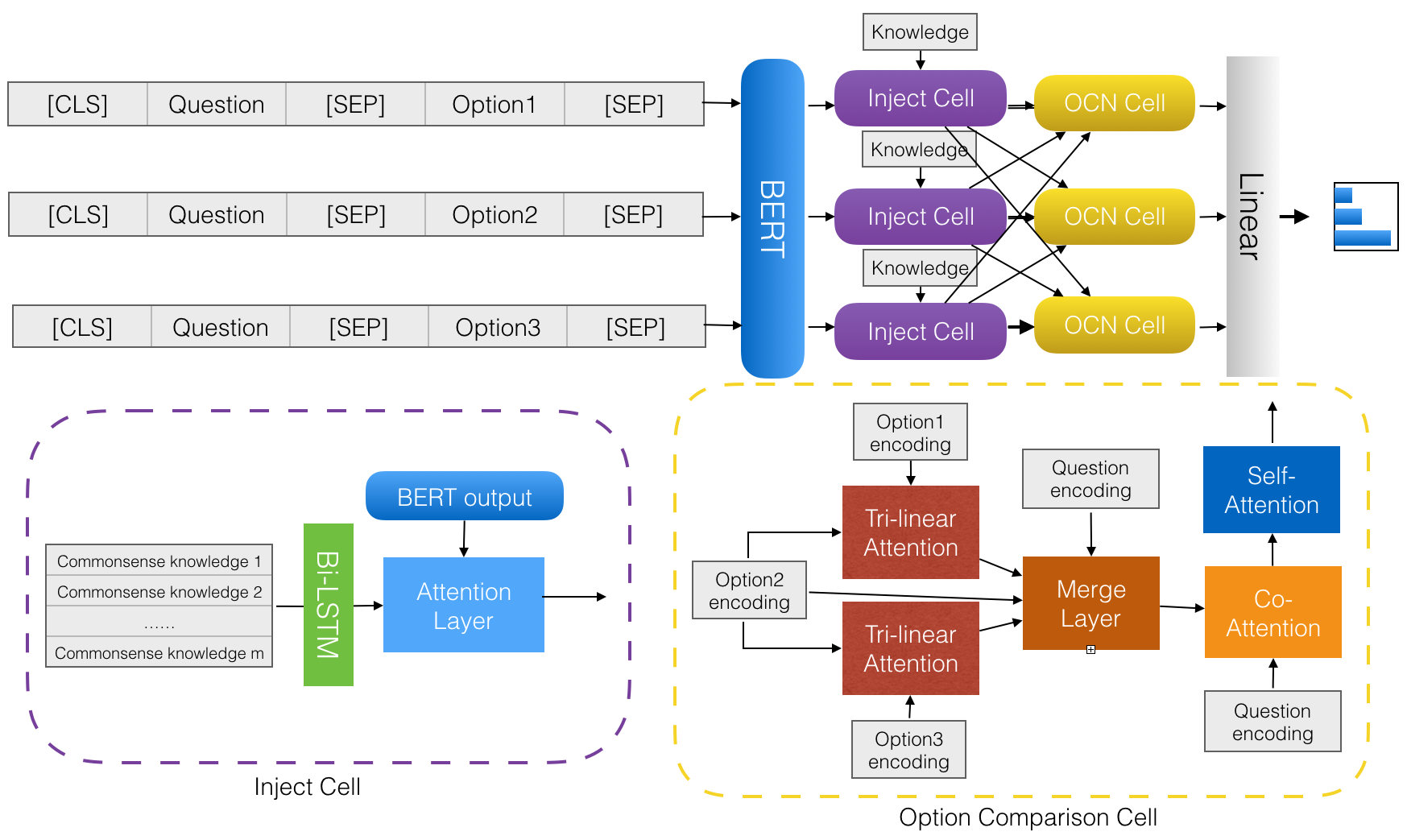}
    \caption{Option Comparison Network (OCN) with Knowledge Injection}
    \label{bosch:fig:model}
\end{figure*}

\subsubsection{Model architecture}

The model class we select is that of the \textit{Bidirectional Encoder Representations with Transformer} (BERT) model \cite{devlin-etal-2019-bert}, as it has been applied to numerous QA tasks and has achieved very promising performance, particularly on the \texttt{CommonsenseQA} dataset. When utilizing BERT on multiple-choice QA tasks, the standard approach is to concatenate the question with each answer-option, in order to generate a list of tokens which is then fed into BERT encoder; a linear layer is added on top, in order to predict the answer. One aspect of this strategy is that each answer-option is encoded independently, which limits the model's ability to find correlations between answer-options and with respect to the original question context. To address this issue, the \textit{Option Comparison Network} (OCN) \cite{DBLP:journals/corr/abs-1903-03033} was introduced to explicitly model the pairwise answer-option interactions, making OCN better-suited for multiple-choice QA task structures. The OCN model uses BERT as its base encoder: the question/option encoding is produced by BERT and further processed in a Option Comparison Cell, before being fed into the linear layer. The Option Comparison Cell is illustrated in the bottom right of figure \ref{bosch:fig:model}. We re-implemented OCN while keeping BERT as its upstream encoder (we refer an interested reader to \cite{DBLP:journals/corr/abs-1903-03033,ma-etal-2019-towards} for more details).

\subsubsection{Knowledge elicitation}
We identify \texttt{ConceptNet} relations that connect questions to the answer-options. The intuition is that these relation paths would provide explicit evidence that would help the model find the answer. Formally, given a question $Q$ and an answer-option $O$, we find all \texttt{ConceptNet} relations \textit{(C1, r, C2)} (where \textit{C1} and \textit{C2} are, respectively, the source and the target node and \textit{r} the relation/edge connecting them), such that $C1 \in Q$ and $C2 \in O$, or vice versa. This rule works well for single-word concepts, but requires some adaptation for the large number of phrases that constitute \texttt{ConceptNet}, in particular relaxing exact-match constraint (for details, see our paper \cite{ma-etal-2019-towards}).\\
We also observe that many questions in the \texttt{CommonsenseQA} task ask about which event is likely to occur, given a pre-condition. Superficially, this particular question type seems well-suited for the procedural nature of \texttt{ATOMIC} triples. However, one challenge of extracting knowledge from this resource is that heads and tails of knowledge triples in \texttt{ATOMIC} are short sentences or verb phrases, while rare words and person-references are reduced to blanks and PersonX/PersonY, respectively.

\subsubsection{Knowledge injection}
The next step is to integrate the extracted triple with the OCN component of our model. Inspired by \cite{bauer-etal-2018-commonsense}, we propose to use attention-based injection. For \texttt{ConceptNet} knowledge triples, we first converted concept-relation entities into tokens from our lexicon, in order to generate a pseudo-sentence. For example, ``\textit{(book, AtLocation, library)}'' would be converted to ``book at location library.'' Next, we used the knowledge injection cell to fuse the commonsense knowledge into BERT's output, before feeding the fused output into the OCN cell. Specifically, in a knowledge-injection cell, a Bi-LSTM layer is used to encode these pseudo-sentences, before computing the attention with respect to BERT output, as illustrated in bottom left of figure \ref{bosch:fig:model}. 

\subsubsection{Knowledge pre-training}
\label{pretrain}
 Our goal is to assess, using \texttt{CommonsenseQA} as benchmark, whether pre-training on \texttt{ConceptNet} and \texttt{ATOMIC} can help the model acquire relevant commonsense to improve performance in specific questions of the task. Our approach follows from a well-known phenomenon studied in NLM: i.e., pre-training large-capacity models (e.g., BERT, GPT \cite{radford2019language}, XLNet \cite{yang2019xlnet}) on large corpora, then fine-tuning on more domain-specific information, leads to increase of accuracy across various tasks. 
 For the \texttt{ConceptNet} pre-training procedure, pre-training BERT on pseudo-sentences formulated from \texttt{ConceptNet} knowledge triples does not provide much gain on performance. Instead, we trained BERT on the \textit{Open Mind Common Sense} (OMCS) corpus \cite{Singh:2002:OMC:646748.701499}, the originating corpus that was used to create \texttt{ConceptNet}. We extracted about 930K English sentences from OMCS and randomly masked out 15\% of the tokens; we then fine-tuned BERT, using a masked language model objective, where the model's objective is to predict the masked tokens, as a probability distribution over the entire lexicon. Finally, we load this fine-tuned model into OCN framework proceed with the downstream \texttt{CommonsenseQA} task. As for pre-training on \texttt{ATOMIC}, we follow previous work's pre-processing steps to convert \texttt{ATOMIC} knowledge triples into sentences \cite{bosselut-etal-2019-comet}; we created special tokens for 9 types of relations as well as blanks. Next, we randomly masked out 15\% of the tokens, only masking out tail-tokens; we used the same OMCS pre-training procedure. 

\begin{table}[h]
\footnotesize
\begin{center}
\caption{Results on \texttt{CommonsenseQA}; the asterisk (*) denotes results taken from leaderboard.}
\label{bosch:csqa-results}
\begin{tabular}{cc}
\toprule \bf Models & \bf Dev Acc  \\ \hline
BERT + OMCS pre-train(*) & 68.8 \\
RoBERTa + CSPT(*) & \bf 76.2 \\ \hline
OCN & 64.1 \\
OCN + CN injection & 67.3 \\
OCN + OMCS pre-train & 65.2 \\
OCN + ATOMIC pre-train & 61.2 \\
OCN + OMCS pre-train + CN inject & \bf 69.0 \\
\bottomrule
\end{tabular}
\end{center}
\end{table}


\subsubsection{Results and Discussion - Part 1}
For all of our experiments, we run 3 trials with different random seeds and we report average scores tables \ref{bosch:csqa-results} and \ref{bosch:csqa-errors1}. Evaluated on \texttt{CommonsenseQA}, \texttt{ConceptNet} knowledge-injection provides a significant performance boost (+2.8\%), compared to the OCN baseline, suggesting that explicit links from question to answer-options help the model find the correct answer. Pre-training on OMCS also provides a small performance boost to the OCN baseline.
Since both \texttt{ConceptNet} knowledge-injection and OMCS pre-training are helpful, we combine both approaches with OCN and we are able to achieve further improvement (+4.9\%). Finally, to our surprise, OCN pre-trained on \texttt{ATOMIC} yields a significantly lower performance.

\begin{table*}[h]
\footnotesize
\begin{center}
\caption{Accuracies for each \texttt{CommonsenseQA} question type: \textbf{AtLoc.} means \textit{AtLocation}, \textbf{Cau.} means Causes, \textbf{Cap.} means \textit{CapableOf}, \textbf{Ant.} means \textit{Antonym}, \textbf{H.Pre.} means \textit{HasPrerequiste}, \textbf{H.Sub} means \textit{HasSubevent}, \textbf{C.Des.} means \textit{CausesDesire}, and \textbf{Des.} means \textit{Desires}. Numbers beside types denote the number of questions of that type.}
\label{bosch:csqa-errors1}
\adjustbox{max width=\textwidth}{
\begin{tabular}{ccccccccc}
\toprule \bf Models & \bf AtLoc.(596) & \bf Cau.(194) & \bf Cap.(109) & \bf Ant.(92) & \bf H.Pre.(46) & \bf H.Sub.(39) & \bf C.Des.(28) & \bf Des.(27) \\ \hline
OCN & 64.9 & 66.5 & 65.1 & 55.4 & 69.6 & 64.1 & 57.1 & 66.7\\
+CN inj, & 67.4(+2.5) & 70.6(+4.1) & 66.1(+1.0) & 60.9(+5.5) & 73.9(+4.3)  & 66.7(+2.6) & 64.3(+7.2) & 77.8(+11.1)\\
+OMCS & 68.8(+3.9) & 63.9(-2.6) & 62.4(-2.7) & 60.9(+5.5) & 71.7(+2.1)  & 59.0(-5.1) & 64.3(+7.2) & 74.1(+7.4)\\
+ATOMIC & 62.8(-2.1) & 66.0(\textbf{-0.5}) & 60.6(-4.5) & 52.2(-3.2) & 63.0(-6.6)  & 56.4(-7.7) & 60.7(\textbf{+3.6}) & 74.1(\textbf{+7.4}) \\
+OMCS+CN & 71.6(+6.7) & 71.6(+5.1) & 64.2(+0.9) & 59.8(+4.4) & 69.6(+0.0)  & 69.2(+5.1) & 75.0(+17.9) & 70.4(+3.7) \\
\bottomrule
\end{tabular}}
\end{center}

\end{table*}

To understand when a model performs better or worse with knowledge-injection, we analyzed predictions by question type. Since all questions in \texttt{CommonsenseQA} require commonsense reasoning, we classify questions based on the \texttt{ConceptNet} relation between the question concept and correct answer concept, \textit{de facto} investigating commonsense \textit{knowledge dimensions} (whose subtle differences have been discussed at length by philosophers, linguists, and cognitive psychologists \cite{davis2014representations}\footnote{Our pragmatic approach, grounded on considering relations/edges in commonsense knowledge graphs (e.g., part\_of, located\_at, cause\_of, etc.) as key knowledge dimensions, led to further work where we specifically looked at the distribution of triples across different resources, studying the impact of each relevant dimension in selected downstream tasks. We demonstrate that identifying the most appropriate commonsense knowledge type(s) required for specific task(s), is instrumental to achieve robust performance. Once the knowledge type is assessed, the most appropriate knowledge resources can be selected, alongside with the corresponding knowledge-extraction methods, and mechanisms to endow language models with relevant triple-based knowledge structures. For details see \cite{ilievski2021dimensions}).}).
The intuition is that the model needs to capture a relevant relation between concepts in order to answer a commonsense question. The accuracy measures for each question type are shown in Table \ref{bosch:csqa-errors1}. Note that the number of samples by question type is very imbalanced. Thus, due to the limited space, we omitted the long tail of the distribution (about 7\% of all samples). We can see that with \texttt{ConceptNet} relation-injection, all question types got performance boosts|for both the OCN model and OCN model that was pre-trained on OMCS | suggesting that external knowledge is indeed helpful for the task. In the case of OCN pre-trained on \texttt{ATOMIC}, although the overall performance is much lower than the OCN baseline, it is interesting to see that performance for the ``Causes'' type is not significantly affected. Moreover, performance for ``CausesDesire'' and ``Desires'' types actually got much better. As noted by \cite{sap2019atomic}, the ``Causes'' relation in \texttt{ConceptNet} is similar to ``Effects'' and ``Reactions'' in \texttt{ATOMIC}; and ``CausesDesire'' in \texttt{ConceptNet} is similar to ``Wants'' in \texttt{ATOMIC}. This result suggests that models with knowledge pre-training perform better on questions that fit the knowledge domain, but perform worse on others. In this case, pre-training on \texttt{ATOMIC} helps the model do better on questions that are similar to \texttt{ATOMIC} relations, even though overall performance is inferior. Finally, we noticed that questions of type ``Antonym'' appear to be the hardest ones: questions that fall into this category typically contain negations, which indicates that horizontal augmentation is not sufficient to endow the model with the ability to reason over negative sentences, a feature that | we hypothesize | requires vertical augmentation. \\
In regards to knowledge-injection methods, attention-based injection seems to be the better choice for pre-trained language models such as BERT: even when alignment between knowledge-base and dataset is sub-optimal, the performance would not degrade. On the other hand, pre-training on knowledge-bases would shift the language model's weight distribution toward its own domain, greatly. If the task domain does not fit knowledge-base well, model performance is likely to drop. When the domain of the knowledge-base aligns with that of the dataset perfectly, both knowledge-injection methods bring performance boosts and a combination of them could bring further gain.


\subsection{Horizontal Augmentation through Knowledge-driven Data Construction for Zero-shot Evaluation}
\label{bosch:sec:zeroshot}

Although what we presented so far provides significant results in support of using symbolic knowledge to augment NLM approaches in Q/A tasks, there is increasing concern that language models overfit to specific datasets, without learning to utilize commonsense knowledge across tasks. 
Zero-shot evaluations have shown promise as a more robust measure of a model's generalizability. While KBs have been shown to help in a zero-shot transfer setting recently ~\cite{banerjee2020self}, no comprehensive study exists on the relation between various knowledge, its usage method, and neural models for zero-shot transfer across commonsense tasks. 
Accordingly, here we show how Q/A data synthetically generated from five knowledge bases  (see section \ref{bosch:sec:kb}) can help to boost performance of language models across the five commonsense question-answering tasks (outlined in \ref{bosch:sec:data}). We found that, while an individual knowledge graph is better suited for specific tasks, a global knowledge graph brings consistent gains across different tasks. We refer the reader to our original work for details on the question generation methods, and on the adversarial filtering techniques \cite{ma2021knowledge}.

\subsubsection{Language Models}
\label{sec:lms}

We consider 2 types of language models: auto-regressive language models and masked language models (MLM). Specifically, we use GPT-2 and RoBERTa to select the best answer candidate. Given a context $C$, a question $Q$, and a list of answer options ($A_1, A_2 ... $), we concatenate $C$ and $Q$ with each answer option to build input sequences ($T_1, T_2 ... $). We also use templates to convert a sequence $T$ into a natural language sentence following \cite{shwartz-etal-2020-unsupervised}. 
For example, we transform the sequence: \textit{[$C$] What will X want to do next? [$A_i$]}  into: \textit{[$C$], as a result, X want to [$A_i$]}. The score $S$ for the resulting sequence using an auto-regressive LM is computed as follows:
\begin{equation}
    \mathrm{S_{LM}}\left(T \right)=-\frac{1}{n} \sum_{i=1}^{n} \log P\left(t_{i} \mid t_{1} \ldots t_{i-1}\right)
\end{equation}
where $n$ is the number of tokens in the sequence and $P$ is the conditional probability provided by the LM. To evaluate MLMs, we mask out one token at a time and compute its loss~\cite{Zhou2020Evaluating}. We repeat this process for every token in the sequence. The final MLM score is:
\begin{equation}
    \mathrm{S_{MLM}}\left(T \right)=-\frac{1}{n} \sum_{i=1}^{n} \log P\left(t_{i} \mid \ldots t_{i-1},  t_{t+1} \ldots \right)
\end{equation}
The predicted option is the one with the lowest score. 

In the typical model architecture for fine-tuning LM for multiple-choice tasks, a linear layer is added on top of the LM encoder to predict the answer. The model inputs are separated by a model-specific delimiter. However, as this architecture introduces randomly initialized parameters, it may not be able to fully utilize the pre-trained weights~\cite{tamborrino2020pre}. 
Instead, we re-use the GPT-2 and RoBERTa with LM head for fine-tuning.
By keeping the model intact, we can reuse the same converting templates and scoring functions.
To train the model, given the scores computed for each answer candidate $S_1, S_2, ... S_m$, we use the marginal ranking (MR) loss defined as:
\begin{equation}
    \mathcal{L}=\frac{1}{m} \sum_{i=1 \atop i \neq y}^{m} \max \left(0, \eta-S_{y}+S_{i}\right)
\end{equation}
Here, $\eta$ represents the margin and $y$ is the index of the correct answer. For a MLM model, the computation cost for the scoring function scales quadratically with the input length. To make the training more efficient, we only mask out non-stop tokens in the head and tail nodes. 



In order to disentangle the contribution of the KGs from the structure of the QA pairs, we consider different training methods for augmentation of language models with KGs. Specifically, we compare marginal ranking (MR) training with masked language modeling (MLM) training. For MLM, we directly concatenate the question and the correct answer in our synthetic QA set and then train RoBERTa on the these sentences using the MLM objective.

\begin{table*}[t]
  \begin{center}
    \caption{
    Zero-shot evaluation results with different combinations of models and knowledge sources, across five commonsense tasks. \texttt{CSKG} represent the combination of \texttt{ATOMIC} and \texttt{CWWV}. We run our experiments three times with different seeds and report average accuracy with 95\% confidence interval. SMLM (*) used OMCS for CSQA, ROCStories \cite{mostafazadeh-etal-2016-corpus} for aNLI and ATOMIC for SIQA as knowledge resources.}
    \label{bosch:tab:zero-shot}
    \resizebox{\linewidth}{!}{
    \begin{tabular}{@{}l@{\hspace{5pt}}c@{\hspace{7pt}}c@{\hspace{7pt}}c@{\hspace{7pt}}c@{\hspace{7pt}}c@{\hspace{7pt}}c@{}} 
     \toprule
    \bf Model & \bf KG &  \bf aNLI  & \bf CSQA & \bf PIQA & \bf SIQA & \bf WG \\
      \midrule
       Majority & - & 50.8 & 20.9 & 50.5 & 33.6 & 50.4 \\
      GPT2-L & - & 56.5  & 41.4 & 68.9 & 44.6 & 53.2 \\
      RoBERTa-L & - & 65.5  & 45.0 &  67.6 & 47.3 & 57.5 \\
      Self-talk \hfill \cite{shwartz-etal-2020-unsupervised} & - & -  & 32.4 & 70.2 & 46.2 & 54.7\\
      COMET-DynaGen \hfill \cite{Bosselut2019DynamicKG} & \texttt{ATOMIC} & - & - & -  & 50.1 & - \\
      SMLM     \hfill \cite{Banerjee2020SelfsupervisedKT} & \texttt{*} & 65.3 & 38.8 & - & 48.5 & - \\
      \midrule
      GPT2-L (MR) & \texttt{ATOMIC} & $59.2(\pm 0.3)$ & $48.0(\pm 0.9)$ & $67.5(\pm 0.7)$ & $53.5(\pm 0.4)$ & $54.7(\pm 0.6)$\\ 
      GPT2-L (MR) & \texttt{CWWV} & $58.3(\pm 0.4)$ & $46.2(\pm 1.0)$ & $68.6(\pm 0.7)$ & $48.0(\pm 0.7)$ & $52.8(\pm 0.9)$\\
      GPT2-L (MR) & \texttt{CSKG} & $59.0(\pm 0.5)$ & $48.6(\pm 1.0)$ & $68.6(\pm 0.9)$ & $53.3(\pm 0.5)$ & $54.1(\pm 0.5)$\\
      RoBERTa-L (MR) & \texttt{ATOMIC} & $\bf 70.8(\pm 1.2)$  &  $64.2(\pm 0.7)$ &  $72.1(\pm 0.5)$ & $63.1(\pm 1.5)$ & $59.6(\pm 0.3)$\\
      RoBERTa-L (MR) & \texttt{CWWV} & $70.0(\pm 0.3)$ &  $\bf 67.9(\pm 0.8)$ & $72.0(\pm 0.7)$  & $54.8(\pm 1.2)$ & $59.4(\pm 0.5)$ \\
      RoBERTa-L (MR) & \texttt{CSKG} & $70.5(\pm 0.2)$ & $67.4(\pm 0.8)$ & $\bf 72.4(\pm 0.4)$ & $\bf 63.2(\pm 0.7)$ & $\bf 60.9(\pm 0.8)$ \\
      \midrule
      \it RoBERTa-L (supervised) & - & 85.6  & 78.5 & 79.2 & 76.6 & 79.3 \\
      \midrule
      \it Human & - & 91.4 & 88.9 & 94.9 & 86.9 & 94.1 \\
    \bottomrule
    \end{tabular}
    }
  \end{center}
\end{table*}

\subsubsection{Experimental Setup}

We compare our results with the following baselines. \textit{Majority} answers each question with the most frequent option in the entire dataset. \textit{`Vanilla' versions of the language models} are used in order to understand the impact of further tuning. Here we directly use the LMs to score the QA pairs without any fine-tuning. We also show the results of other unsupervised systems that leverage KGs: \textit{Self-talk}, \textit{COMET-}\textit{DynaGen}, and \textit{SMLM}. To indicate the upper bound of this work, we include results of a supervised fine-tuned RoBERTa system and of human evaluation. 

\subsubsection{Results and Discussion - Part 2}
Table 4 shows that GPT-2 and RoBERTa outperform the majority baseline by a large margin on all tasks, indicating that the LMs have already learned relevant knowledge during pre-training. Despite being a smaller model, RoBERTa outperforms GPT-2 on 4 out of 5 tasks without pre-training, and on all tasks when pre-training over different synthetic QA sets. This shows the advantage of leveraging bi-directional context. Training RoBERTa on our \texttt{ATOMIC} or \texttt{CWWV} synthetic sets brings notable performance gain on all 5 tasks. We observe that models trained on \texttt{ATOMIC} sets have a large advantage on SIQA compare to models trained on \texttt{CWWV}, while \texttt{CWWV} brings advantage on the CSQA task. This is not surprising as these two tasks are derived from \texttt{ConceptNet} and \texttt{ATOMIC}, respectively. The difference between \texttt{ATOMIC} and \texttt{CWWV} on the remaining three tasks is relatively small. This supports the main findings presented in section \ref{bosch:sec:attention}: knowledge alignment is crucial for obtaining better performance. Training on the combined question set (\texttt{CSKG}) is mostly able to retain the best of its both partitions. Training on \texttt{CSKG} leads to best performance on three out of five tasks, showing that a global commonsense resource is able to bring consistent gain across different tasks. This shows that adding more diverse knowledge is beneficial for language models. The training regime had a role too: marginal ranking is superior to vanilla language modeling, as it preserved better the structure of the task. Future work should also investigate the impact of this approach on knowledge injection systems, as for instance our attention-based model presented in section \ref{bosch:sec:attention}.


\subsection{Vertical Augmentation: Towards Exploring the Depth of Common Sense Reasoning}
\label{bosch:sec:depth}

In past work \cite{prevot2005interfacing,huang2010ontology}, we observed that it's a signature characteristic of computational lexical resources to have high \textit{concept density}, namely large number of interconnected classes, but low \textit{constraint density}, that is a limited number of axioms used to formalize the semantic relations holding between nodes. Although this distinction was made before the advent of Big Data and the consequent explosion of machine learning-inspired AI, the underlying observation remains valid today, if not even more relevant, when applied to the current landscape of lexical/semantic resources: in particular, when we consider commonsense knowledge graphs, the attention is mostly on size and not on the level of formalization. The explanation is manifold: 1) since the vast majority of resources is based on triple-like structures, the constraint options are bound to the expressivity of the RDF language \cite{pan2009resource}; 2) crowd-sourcing played a key role in building  many seminal resources like \texttt{ConceptNet}, which implies that complying to strict formalization rules was not a design requirement; 3) deep learning represents the mainstream approach to solve a plethora of downstream tasks, as the popularity of language models in NLP testifies, and only very large-scale knowledge resources can be leveraged by deep neural networks\footnote{One of the main reasons why research on machine common sense reasoning is experiencing a \textit{renaissance} period in AI is thanks to initiatives like the homonymous DARPA program \cite{gunning2018machine}, which promotes the integration between symbolic knowledge and sub-symbolic (neural) algorithms.}. \\
The body of work presented in the previous sections, focused on horizontal augmentation of language models, can be conceived as an example of what was outlined in point 3) above: we exploited the massive scale of general facts available in state-the-art knowledge graphs, in orchestration with most advanced NLM approaches. However, as claimed in section \ref{bosch:sec:lack}, improving models performance in Q/A tasks also depends on enabling specific forms of knowledge-based reasoning (\textbf{vertical augmentation}): thus, introducing fine-grained constraints in knowledge graphs becomes an essential step towards robustness and generalizability. But how can this enhancement be implemented if, as stated in point 1), we are dealing with semantic frameworks defined in terms of \textit{subject/predicate/object} primitives? \\

\subsubsection{Enhancing the coverage of spatial-commonsense knowledge}

In a recent publication \cite{ilievski2021dimensions}, we identified 13 dimensions shared by the most relevant common sense knowledge graphs in the state of the art (see section \ref{bosch:sec:kb}). In particular, our study shows that the temporal and intentional dimensions are beneficial for reasoning on Q/A downstream tasks, while lexical information have little impact.\footnote{For more details we refer the reader to the original article.} In some cases, such as the spatial dimension, the relations used by knowledge graphs are too generic to help in the downstream tasks. For instance, \texttt{ConceptNet} realizes the spatial dimension through the \textit{AtLocation} edge: when we consider spatially-relevant questions from the dataset presented in section \ref{bosch:sec:data}, knowledge about the generic location of entities is clearly insufficient to answer questions that require reasoning over position, orientation, motion.  As an example, in \texttt{PIQA}, no language model can answer the question `how do you wear a shawl', with candidates `place it \textit{on} your shoulders' and `place it \textit{under} your shoulders', just by acquiring a triple  `shawl \textit{AtLocation} shoulders' from ConceptNet. It is irrelevant whether we use an attention-based injection method (\ref{bosch:sec:attention}), a model pre-trained on synthetically-generated questions and answers (\ref{bosch:sec:zeroshot}), or an alternative approach: models would need to be provided with a triple that includes the specific spatial relation \textit{on}, connecting the concept of `shawl' with the concept of `shoulders', or some related inference.\footnote{For instance, a model can be provided with information that jackets are typically worn on shoulders, and learn to generalize that garments which share similar characteristics, like shawls, should also be worn on the shoulders, and not under them (like, for instance, corsets).}
\texttt{Visual Genome}, one of the federated resources included in \texttt{CSKG} (see section \ref{bosch:sec:kb}), can help to combat this problem, as it supports a rich set of specific spatial relations. In particular, we propose the following procedure to perform \textbf{vertical augmentation} of language models in spatial commonsense Q/A tasks using \texttt{Visual Genome}:



\begin{itemize}
    \item \textbf{Step 1} Extract triples that include spatial relations from \texttt{Visual Genome}. We only consider triples that appear more than three times in the whole dataset as common knowledge (following the same method used in \cite{yang2018visual}). 

    \item \textbf{Step 2} Generate synthetic questions and answers for each triple, and pre-train RoBERTa-Large and GPT-2, following the method mentioned in section \ref{bosch:sec:zeroshot}. Each sentence must have all the components of the triple in the correct order (i.e., [HEAD]|[RELATION]|[TAIL]), to be considered as a valid linguistic representation of the triple. In general, for generating questions, we convert each sentence to a question by masking the relevant spatial relation (Ex. \textit{Bike parked near the trees} is converted to \textit{Bike parked [MASK] the trees}.) 

    \item \textbf{Step 3} To pre-train the language models in an adequate manner, we need to have fair and challenging answer options, i.e. candidates that exemplify which spatial relations can exist between two entities and which cannot. To this end, we categorized all spatial relations in \texttt{Visual Genome} into spatial classes (for instance, \textit{above, over, up, top, overhead, north, upside} are classified using the category \textit{ABOVE}). These classes help us to consolidate triples on the basis of their most salient spatial information. For instance, for the question ``The pedestrian [MASK] the street'', [Pedestrian, On, Street], [Pedestrian, Above, Street] are unified using ABOVE as spatial category. 

    \item \textbf{Step 4} Further pre-train RoBERTa-Large and GPT-2 (already pre-trained on CSKG) on the spatial Q/A data synthetically generated as described in steps 1-3. 

    \item \textbf{Step 5} Select a subset of spatially-relevant Q/A pairs from the commonsense benchmarks, using a combination of extraction techniques, including a constituency parser and OIE (Open Information Extraction) models \cite{angeli2015leveraging}. Table \ref{bosch:spatial_questions} provides the number of extracted examples from the Dev set of each benchmark.
    
    \item \textbf{Step 6} Evaluate performance both globally, i.e., for the whole dataset under consideration, and locally, i.e., for the partitions that contain only spatially relevant questions. 

\end{itemize} 

\begin{table}[h]
\begin{center}
\caption{Number of examples with spatial information from different benchmarks}
\label{bosch:spatial_questions}
\adjustbox{max width=\textwidth}{
\begin{tabular}{ccc}
\toprule \bf Dataset  & \bf  Dev. size& \bf Extracted spatial examples \\ 
\hline
CommonsenseQA~\cite{talmor-etal-2019-commonsenseqa} & 1221 & 800\\
PhysicalIQA~\cite{Bisk2020} & 1838 & 1209\\
SocialIQA~\cite{sap-etal-2019-social} & 1954 & 1142\\
WinoGrande~\cite{sakaguchi2019winogrande} & 1267 & 431\\
aNLI~\cite{bhagavatula2019abductive} & 1532 & 1077 \\ \bottomrule
\end{tabular}}
\end{center}
\end{table}

Regarding the evaluation of the approach (\textbf{Step 6}), two observations can be made at the time of this publication: preliminary results show that 1) pre-training language models on specific spatial knowledge does not alter the general commonsense knowledge gained from pre-training on \texttt{CSKG}; 2) a slight improvement (+2\%) is observed only for the SIQA dataset, which indicates that the knowledge extracted from \texttt{Visual Genome} is substantially misaligned with the spatial domain partition of the other benchmarks (a phenomenon also reported in \cite{ma2020knowledgedriven, mirzaee2021spartqa}). Due to the early stage of this research work on vertical augmentation, a thorough error analysis cannot be reported here. However, limited random sampling seem to corroborate the anticipated benefits of our approach. For instance, spatial knowledge from \texttt{Visual Genome} helped the model to correctly answer the \texttt{PIQA} question \textit{How do you make a plain toast?}, with [Option 1]: ``Gather a slice of bread and a toaster, place the bread \textit{on the top of the toaster} and push the button down"; and [Option 2]:  ``Gather a slice of bread and a toaster, place the bread \textit{into the toaster} and push the button down".\\
The procedure described in this section exploits pre-existing information in the \texttt{Visual Genome} resource and does not prescribe any structural modification to the spatial triples of \texttt{CSKG}. However, the spatial classes mentioned in \textbf{Step 3} could be used to upgrade the spatial relations currently implemented in \texttt{CSKG}. Furthermore, we hypothesize that, by modifying \texttt{Visual Genome} through additional annotations based on formally-characterized spatial relations, reasoning capabilities can be advanced further. In particular, as demonstrated by \cite{mirzaee2021spartqa}, it is possible to obtain high quality textual annotations of topological relations like \textit{connected}, \textit{disconnected}, \textit{overlap}, \textit{in} and \textit{touch}\footnote{See \cite{randell1992spatial, renz2007qualitative} for an overview of qualitative spatial reasoning, and \cite{kordjamshidi2010spatial} for an introduction to methods of extraction of spatial relations from text.}, and boost the performance of models in datasets like bAbI \cite{weston2015towards} (see section \ref{bosch:sec:lack}).

%

\section{Conclusions}

As much as the `capability to generalize' has been considered, for long time, one of the requisites that an artificial system should satisfy in order to exhibit `intelligence' \cite{simon1990bounded}, building generalizable systems represents all but a solved problem in today's AI research. In this chapter, we demonstrated that, within the scope of commonsense question answering,  neural language models can achieve robustness within and across downstream tasks, when integrated with large-scale knowledge graphs, according to mechanisms like attention-based injection and pre-training with synthetic questions and answers generated from knowledge graphs. We showed that, through \textbf{horizontal augmentation}, language models can access a vast trove of knowledge in the form of triples. Finally, we made the case for enhancing reasoning capabilities in neuro-symbolic systems by introducing more fine-grained semantic relations in the knowledge graphs they are combined with (\textbf{vertical augmentation}), outlining a procedure to test this hypothesis. Supported by our empirical investigations, and inspired by recent literature in integrating qualitative reasoning with neural models, we have provided the blueprints of a solution towards constructing generalizable neuro-symbolic systems for commonsense question answering. 

\section{Acknowledgements}
The authors would like to thank you all the students and faculty who contributed, with their hard work and precious advice, to make this chapter possible: Monireh Ebrahimi, Sarah Masud Preum, Bin Zhang, Eric Nyberg, Pedro Szekely, Deborah L. McGuinness, and Yonatan Bisk.

\bibliographystyle{tfnlm}
\bibliography{chap}

\end{document}